\documentclass{article}
\usepackage{spconf,amsmath,graphicx}
\usepackage{booktabs, multirow}
\usepackage{tikz} 
\usepackage{svg}
\usepackage{amsmath}
\usepackage{hyperref}

\title{TUNet: A block-online bandwidth extension model based on Transformers and self-supervised pretraining}

\name{Viet-Anh Nguyen$^{1}$, Anh H. T. Nguyen$^{1}$, and Andy W. H. Khong$^{2}$}
\address{$^{1}$NextG, FPT Software, Vietnam\\
$^{2}$Nanyang Technological University, Singapore\\
\texttt{\{anhnv79, anhnht3\}@fsoft.com.vn, andykhong@ntu.edu.sg
}}

\begin{document}
%

\maketitle

\begin{abstract}
We introduce a block-online variant of the temporal feature-wise linear modulation (TFiLM) model to achieve bandwidth extension. The proposed architecture simplifies the UNet backbone of the TFiLM to reduce inference time and employs an efficient transformer at the bottleneck to alleviate performance degradation. We also utilize self-supervised pretraining and data augmentation to enhance the quality of bandwidth extended signals and reduce the sensitivity with respect to downsampling methods. Experiment results on the VCTK dataset show that the proposed method outperforms several recent baselines in both intrusive and non-intrusive metrics. Pretraining and filter augmentation also help stabilize and enhance the overall performance.
\end{abstract}
\vspace{-0.2cm}
\section{Introduction}
\vspace{-0.1cm}
\label{sec:intro}
Bandwidth extension (BWE), or audio super-resolution, enhances speech by generating a wideband (WB) signal from a narrowband (NB) signal. The NB signal is usually sampled below 8 kHz resulting in low auditory quality. Such sampling rate is widely used in G.711, G.729, and AMR audio codecs due to its efficient streaming. Including a BWE module at the receiver side will therefore improve audio fidelity.

Compared to conventional BWE approaches such as~\cite{codebook, gmm2, hmm}, recent end-to-end deep neural networks generate WB signals directly from NB signals without the need for feature engineering. For instance, inspired by the well-known UNet architecture \cite{ronneberger2015unet} in image processing, AudioUNet \cite{waveunet} is a wave-to-wave BWE model that has outperformed traditional methods. In \cite{birnbaum2021temporal}, the limitation of convolution on long-range dependency modeling in UNet is addressed by introducing the TFiLM layer that modulates blocks of convolution's feature maps with information learned by recurrent layers. Generative models such as the NU-Wave \cite{nuwave} neural vocoder relies on conditional diffusion models with modified noise level embedding and local conditioner. On the other hand, WSRGlow \cite{wsrglow} models the distribution of the output conditioned on the input using normalizing flow.

\begin{figure*}[htb]
\centering
\tikzset{every picture/.style={line width=0.75pt}} 

\tikzset{every picture/.style={line width=0.75pt}} 

\tikzset{every picture/.style={line width=0.75pt}} 
\resizebox{13cm}{3cm}{%
\begin{tikzpicture}[x=0.85pt,y=0.60pt,yscale=-1,xscale=1]

\draw (105.03,288.3) node [rotate=-270] {\includegraphics[width=94.3pt,height=20.51pt]{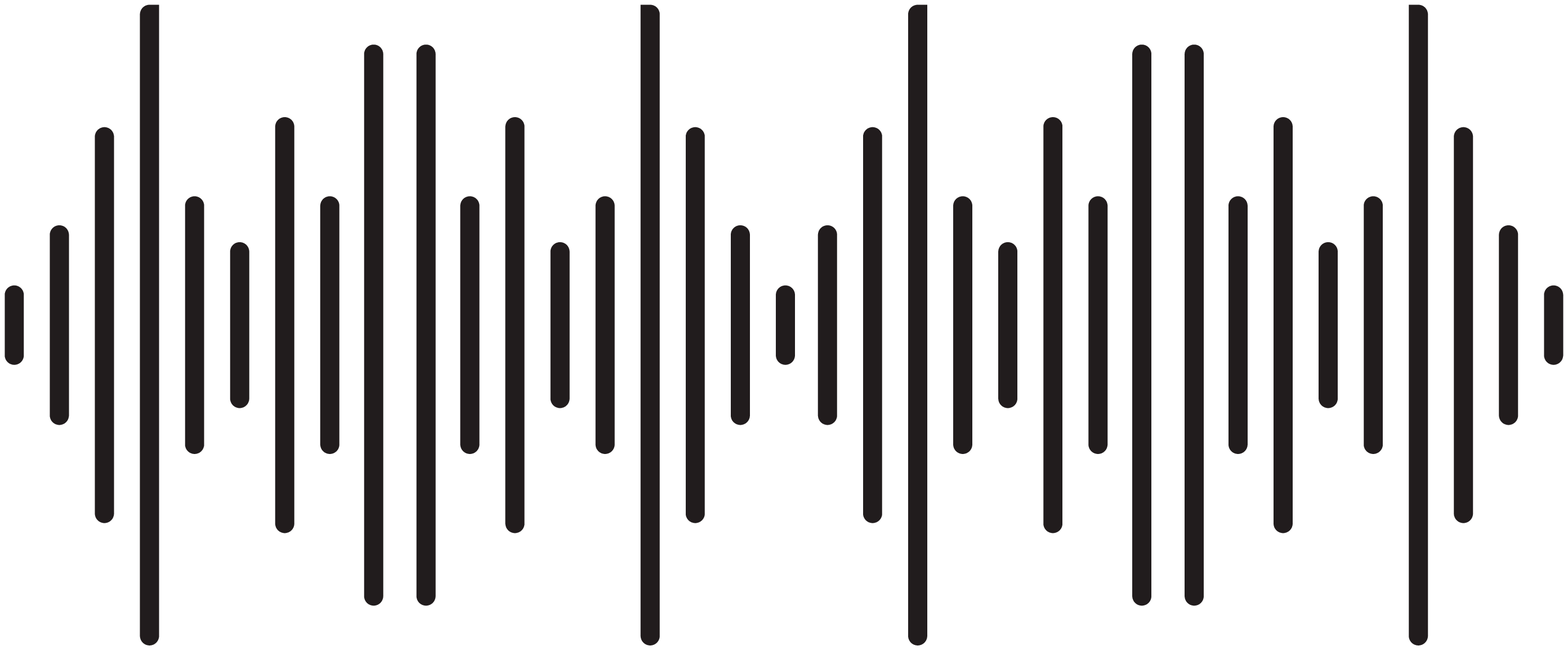}};
\draw  [fill={rgb, 255:red, 74; green, 144; blue, 226 }  ,fill opacity=1 ] (148.54,195.5) -- (168.99,210.79) -- (169.39,362.74) -- (149.02,383.12) -- cycle ;
\draw  [fill={rgb, 255:red, 179; green, 197; blue, 218 }  ,fill opacity=1 ] (186.77,356.73) -- (186.42,215.54) -- (208.76,215.54) -- (209.11,356.73) -- cycle ;
\draw  [fill={rgb, 255:red, 74; green, 144; blue, 226 }  ,fill opacity=1 ] (225.88,216.61) -- (246.32,222.04) -- (246.6,350.94) -- (226.22,362.45) -- cycle ;
\draw  [fill={rgb, 255:red, 179; green, 197; blue, 218 }  ,fill opacity=1 ] (261.84,351.72) -- (261.58,227.39) -- (283.92,227.34) -- (284.18,351.67) -- cycle ;
\draw  [fill={rgb, 255:red, 74; green, 144; blue, 226 }  ,fill opacity=1 ] (300.04,226.54) -- (320.46,236.23) -- (320.68,339.54) -- (300.29,349.32) -- cycle ;
\draw [line width=1.5]    (169.19,288.69) -- (181.65,288.66) ;
\draw [shift={(184.65,288.65)}, rotate = 539.88] [color={rgb, 255:red, 0; green, 0; blue, 0 }  ][line width=1.5]    (8.53,-2.57) .. controls (5.42,-1.09) and (2.58,-0.23) .. (0,0) .. controls (2.58,0.23) and (5.42,1.09) .. (8.53,2.57)   ;
\draw [line width=1.5]    (209.41,289.25) -- (222.11,289.09) ;
\draw [shift={(225.11,289.05)}, rotate = 539.29] [color={rgb, 255:red, 0; green, 0; blue, 0 }  ][line width=1.5]    (8.53,-2.57) .. controls (5.42,-1.09) and (2.58,-0.23) .. (0,0) .. controls (2.58,0.23) and (5.42,1.09) .. (8.53,2.57)   ;
\draw [line width=1.5]    (246.29,287.96) -- (257.55,288) ;
\draw [shift={(260.55,288.01)}, rotate = 180.2] [color={rgb, 255:red, 0; green, 0; blue, 0 }  ][line width=1.5]    (8.53,-2.57) .. controls (5.42,-1.09) and (2.58,-0.23) .. (0,0) .. controls (2.58,0.23) and (5.42,1.09) .. (8.53,2.57)   ;
\draw [line width=1.5]    (284.12,288.61) -- (297.3,288.45) ;
\draw [shift={(300.3,288.41)}, rotate = 539.31] [color={rgb, 255:red, 0; green, 0; blue, 0 }  ][line width=1.5]    (8.53,-2.57) .. controls (5.42,-1.09) and (2.58,-0.23) .. (0,0) .. controls (2.58,0.23) and (5.42,1.09) .. (8.53,2.57)   ;
\draw [line width=1.5]    (131.24,289.25) -- (145.13,289.22) ;
\draw [shift={(148.13,289.21)}, rotate = 539.88] [color={rgb, 255:red, 0; green, 0; blue, 0 }  ][line width=1.5]    (8.53,-2.57) .. controls (5.42,-1.09) and (2.58,-0.23) .. (0,0) .. controls (2.58,0.23) and (5.42,1.09) .. (8.53,2.57)   ;
\draw  [fill={rgb, 255:red, 224; green, 255; blue, 193 }  ,fill opacity=1 ] (341.52,337.27) .. controls (338.75,337.27) and (336.5,335.03) .. (336.5,332.26) -- (336.31,244.02) .. controls (336.3,241.25) and (338.54,239) .. (341.31,239) -- (359.93,238.96) .. controls (362.7,238.95) and (364.95,241.19) .. (364.96,243.96) -- (365.14,332.2) .. controls (365.15,334.97) and (362.91,337.22) .. (360.14,337.23) -- cycle ;
\draw [line width=1.5]    (365.13,288.35) -- (379.29,288.26) ;
\draw [shift={(382.29,288.24)}, rotate = 539.65] [color={rgb, 255:red, 0; green, 0; blue, 0 }  ][line width=1.5]    (8.53,-2.57) .. controls (5.42,-1.09) and (2.58,-0.23) .. (0,0) .. controls (2.58,0.23) and (5.42,1.09) .. (8.53,2.57)   ;
\draw [line width=1.5]    (320.29,288.05) -- (332.75,288.28) ;
\draw [shift={(335.75,288.34)}, rotate = 181.07] [color={rgb, 255:red, 0; green, 0; blue, 0 }  ][line width=1.5]    (8.53,-2.57) .. controls (5.42,-1.09) and (2.58,-0.23) .. (0,0) .. controls (2.58,0.23) and (5.42,1.09) .. (8.53,2.57)   ;
\draw  [fill={rgb, 255:red, 74; green, 144; blue, 226 }  ,fill opacity=1 ] (551.28,382.67) -- (530.86,370.9) -- (530.9,206.74) -- (551.32,195.05) -- cycle ;
\draw  [fill={rgb, 255:red, 179; green, 197; blue, 218 }  ,fill opacity=1 ] (513.51,205.68) -- (513.48,372.48) -- (491.12,372.43) -- (491.15,205.63) -- cycle ;
\draw  [fill={rgb, 255:red, 74; green, 144; blue, 226 }  ,fill opacity=1 ] (474.02,372.19) -- (453.6,360.13) -- (453.63,218.34) -- (474.05,206.36) -- cycle ;
\draw  [fill={rgb, 255:red, 179; green, 197; blue, 218 }  ,fill opacity=1 ] (438.39,222.82) -- (438.36,354.78) -- (416.01,354.73) -- (416.04,222.78) -- cycle ;
\draw [line width=1.5]    (399.15,288.27) -- (413.31,288.19) ;
\draw [shift={(416.31,288.17)}, rotate = 539.65] [color={rgb, 255:red, 0; green, 0; blue, 0 }  ][line width=1.5]    (8.53,-2.57) .. controls (5.42,-1.09) and (2.58,-0.23) .. (0,0) .. controls (2.58,0.23) and (5.42,1.09) .. (8.53,2.57)   ;
\draw [line width=1.5]    (438.18,288.67) -- (451.26,288.59) ;
\draw [shift={(454.26,288.57)}, rotate = 539.63] [color={rgb, 255:red, 0; green, 0; blue, 0 }  ][line width=1.5]    (8.53,-2.57) .. controls (5.42,-1.09) and (2.58,-0.23) .. (0,0) .. controls (2.58,0.23) and (5.42,1.09) .. (8.53,2.57)   ;
\draw [line width=1.5]    (473.98,289.08) -- (488.14,289) ;
\draw [shift={(491.14,288.98)}, rotate = 539.65] [color={rgb, 255:red, 0; green, 0; blue, 0 }  ][line width=1.5]    (8.53,-2.57) .. controls (5.42,-1.09) and (2.58,-0.23) .. (0,0) .. controls (2.58,0.23) and (5.42,1.09) .. (8.53,2.57)   ;
\draw [line width=1.5]    (514.24,289) -- (528.4,288.91) ;
\draw [shift={(531.4,288.89)}, rotate = 539.65] [color={rgb, 255:red, 0; green, 0; blue, 0 }  ][line width=1.5]    (8.53,-2.57) .. controls (5.42,-1.09) and (2.58,-0.23) .. (0,0) .. controls (2.58,0.23) and (5.42,1.09) .. (8.53,2.57)   ;
\draw [line width=1.5]    (551.83,288.44) -- (565.99,288.35) ;
\draw [shift={(568.99,288.33)}, rotate = 539.65] [color={rgb, 255:red, 0; green, 0; blue, 0 }  ][line width=1.5]    (8.53,-2.57) .. controls (5.42,-1.09) and (2.58,-0.23) .. (0,0) .. controls (2.58,0.23) and (5.42,1.09) .. (8.53,2.57)   ;
\draw (584.51,289.76) node [rotate=-270] {\includegraphics[width=94.75pt,height=19.27pt]{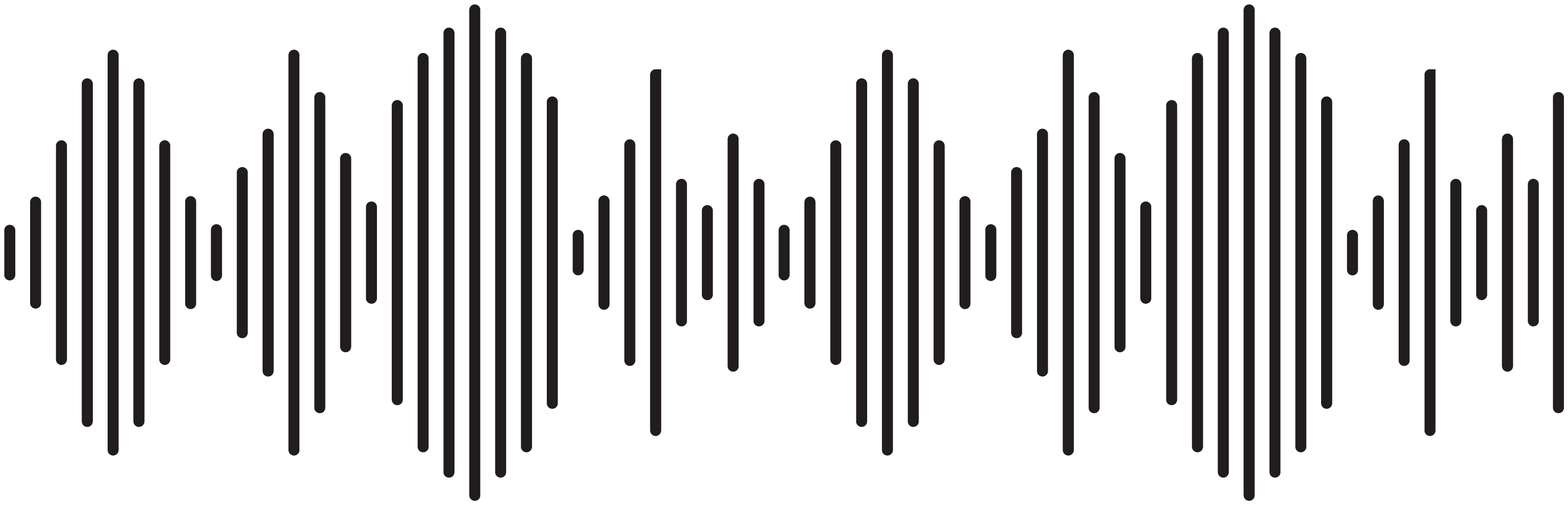}};
\draw [line width=1.5]    (328,288.19) -- (328,225) -- (372,225) -- (372,288.42) ;
\draw [line width=1.5]    (292,288.51) -- (292,215) -- (444,215) -- (444,288.45) ;
\draw [line width=1.5]    (213,289.08) -- (213,197) -- (523,197) -- (523,288.95) ;
\draw [line width=1.5]    (140,289.23) -- (140,182) -- (558.5,182) -- (558.1,287.39) ;
\draw  [fill={rgb, 255:red, 74; green, 144; blue, 226 }  ,fill opacity=1 ] (401.99,352.97) -- (381.56,342.2) -- (381.33,234.48) -- (401.72,223.62) -- cycle ;

\draw (75.03,288.3) node  [font=\normalsize,rotate=-270] [align=left] {{\normalsize $L_{in} = 8192$}};

\draw (351.81,289.06) node  [font=\normalsize,rotate=-270] [align=left] {{\footnotesize Transformer}};
\draw (392.92,287.3) node  [font=\normalsize,rotate=-270] [align=left] {{\footnotesize TConv(C=128, K=8) }};
\draw (311.89,286.29) node  [font=\normalsize,rotate=-270] [align=left] {{\footnotesize Conv(C=256, K=8) }};
\draw (273.94,288.23) node  [font=\normalsize,rotate=-270] [align=left] {{\footnotesize TFiLM(B=64)}};
\draw (199.85,289.15) node  [font=\normalsize,rotate=-270] [align=left] {{\footnotesize TFiLM(B=64)}};
\draw (427.67,288.57) node  [font=\normalsize,rotate=-270] [align=left] {{\footnotesize TFiLM(B=64)}};
\draw (502.58,288.91) node  [font=\normalsize,rotate=-270] [align=left] {{\footnotesize TFiLM(B=64)}};
\draw (464.58,288.82) node  [font=\normalsize,rotate=-270] [align=left] {{\footnotesize TConv(C=64, K=18) }};
\draw (541.96,287.31) node  [font=\normalsize,rotate=-270] [align=left] {{\footnotesize TConv(C=1, K=66) }};
\draw (236.69,283.17) node  [font=\normalsize,rotate=-270] [align=left] {{\footnotesize Conv(C=128, K=18) }};
\draw (159.79,287.98) node  [font=\normalsize,rotate=-270] [align=left] {{\footnotesize Conv(C=64, K=66) }};
\draw (610.51,289.76) node  [font=\normalsize,rotate=-270] [align=left] {{\normalsize $L_{out} = 8192$}};

\end{tikzpicture}
}

\caption{TUNet architecture for speech enhancement. The encoder downsamples waveform input while the decoder does the reverse. A Transformer block is placed in the middle to model the attention of the bottleneck.}
\label{fig:unet}
\end{figure*}

While convolutional neural network architectures exhibit promising results for end-to-end BWE training, their effectiveness on long-range dependency modeling is still limited by receptive fields of convolution \cite{10.5555/3326943.3326958}. Stacking more convolution layers would help expand the receptive field at the expense of increased computation. In addition, training end-to-end BWE models requires high-rate target signals, making valuable low-rate data collected from telephony 8-kHz infrastructure unusable. 
It has also been observed that BWE models are susceptible to low-pass filtering \cite{birnbaum2021temporal, Sulun_2021}, generating severe distortion at the transition band of the anti-aliasing filter. This problem can be mitigated by data augmentation \cite{Sulun_2021}.


We propose a Transformer-aided UNet (TUNet)\footnote{Source code and audio samples: \url{https://github.com/NXTProduct/TUNet}} by employing a low-complexity transformer encoder on the bottleneck of a lightweight UNet. Here, the Transformer assists such a small UNet with its captured global dependency while the UNet effectively downsamples waveform input with strided convolution to reduce computation that the Transformer must perform. In addition, inspired by masked language modeling in natural language processing \cite{devlin2019bert}, we propose \textit{masked speech modeling} --- a self-supervised representation learning scheme that reconstructs original signals from masked signals. The advantage of this pretraining is that it requires only low-rate data to make full use of telephony databases, allowing the model to learn the underlying statistics of the low-band speech and generalize to downstream tasks \cite{ssup1}. Finally, similar to \cite{Sulun_2021}, we make our model robust to downsampling methods by generating training data with different parameter sets of the Chebyshev Type I filter. \vspace{-0.2cm}


\section{Review of TFiLM-UNet and proposed TUNet algorithm  }
\vspace{-0.2cm}
\subsection{TFiLM-UNet baseline}
TFiLM-UNet is an offline UNet-based audio super-resolution model \cite{birnbaum2021temporal}. To assist convolution layers in capturing long-range information, Temporal Feature-Wise Linear Modulation (TFiLM) has been proposed. This layer acts as a normalization layer that combines maxpooling and long short-term memory (LSTM). While maxpooling reduces temporal dimension into $B$ blocks, LSTMs refine convolution's feature maps by captured long-range dependency.

In the TFiLM-UNet model, the encoder contains four downsampling (D) blocks, each comprising a convolution layer, maxpooling layer, ReLU activation, and TFiLM layer, consecutively. In the decoder, upsampling (U) blocks follow sequential operations: convolution, dropout, ReLU, DimShuffle, and TFiLM, in which the DimShuffle layer doubles time dimension by manipulating the feature shape. Stacking and additive skip connections are applied between D/U blocks and input/output, respectively.
\vspace{-0.2cm}
\subsection{Lightweight UNet with Transformer}

With reference to Fig.~\ref{fig:unet}, our proposed model follows the same waveform-to-waveform UNet to that of TFiLM. As opposed to TFiLM-UNet, the proposed model is significantly smaller due to the use of fewer convolution filters and higher dimensional reduction rates. Precisely, the encoder consists of three strided 1D convolution layers, each having $C$ filters of kernel size $K$. Stride $S$ of all these layers is set at 4, resulting in the time dimension of the bottleneck being 64 times shorter than the length $L_{in}$ of the input. Consequently, the bottleneck features can be processed efficiently in the follow-up Performers \cite{choromanski2021rethinking} blocks. We employ Performers since its self-attention mechanism has linear time complexity compared to the quadratic complexity of the conventional attention \cite{vaswani2017attention}. On the decoder side, three transposed 1D convolution layers commensurating the downsampling rates of the encoder are used to generate output signals that have the length $L_{out}=L_{in}$. We use Tanh activation for the last transposed convolution and LeakyReLU \cite{Maas2013RectifierNI} for the rest. TFiLM layers are applied after convolution layers except for the last encoder layer that is replaced by the Performer blocks. To smooth the loss landscape~\cite{Wang2020IsTS}, skip connections that connect TFiLM encoders to the corresponding decoders are employed.

Compared to the TFiLM-UNet, our model has four key differences: i) Our encoder and decoder require one fewer layer and four times fewer filters than the baseline; ii) Each encoding layer reduces time dimension by four times instead of two to assist quick input compression; iii) We replace downsampling and upsampling blocks in TFiLM with strided convolution and transposed convolution layers, respectively; and iv) compared to the stacking skip-connection in TFiLM, we employ additive skip connection which further reduces the number of parameters in the decoder. These modifications ensure that our model is significantly lighter than the baseline while preserving learning capability.\vspace{-0.1cm}

\vspace{-0.2cm}
\subsection{Masked speech modeling}
\label{sec:msm}
\begin{figure}
    \centering
    \includegraphics[width=7.5cm]{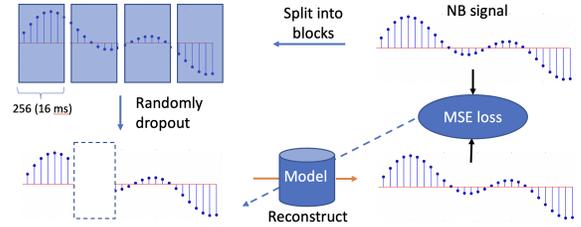}
    \caption{Masked speech modeling pretraining pipeline.}
    \label{fig:msm}
\end{figure}
We propose masked speech modeling (MSM) pretraining as illustrated in Fig.~\ref{fig:msm}. Since audio signals possess fine granular characteristics, instead of masking the sequence at sample level, we mask 20\% of 256-sample blocks to create the masked input. The model will optimize the mean squared error between the output and the masked input. Compared to the masked reconstruction pretraining in~\cite{wang2020unsupervised}, both encoder and decoder are pretrained in our proposed approach.
\vspace{-0.2cm}

\subsection{Improving robustness to downsampling methods by augmentation}
\label{sec:mfd}
The performance of BWE models is highly sensitive to different anti-aliasing filters when downsampling methods in testing differ from training \cite{waveunet, birnbaum2021temporal, Sulun_2021}. Similar to \cite{Sulun_2021}, to improve the robustness of our model, we generate the low-rate signals by downsampling the high-rate speech dataset with random anti-aliasing filters. More specifically, we adopt the Chebyshev Type I anti-aliasing filter and randomize its ripple and order parameters. This helps in creating variations in the transition band of the anti-aliasing filter.\vspace{-0.1cm}

\subsection{Learning objectives}
Since the mean squared error (MSE) loss may not guarantee the good perceptual quality \cite{perceptual2}, we combine MSE loss with multi-resolution short-time Fourier transform (STFT) loss \cite{mulres} in the Mel scale. Given a reconstructed signal $\hat{y}$ and a target signal $y$, the training loss is given by \vspace{-0.3cm}

\begin{equation}
\label{eq:loss}
\ell(\hat{y}, y)=\ell_{\mathrm{MR}}(\hat{y}, y) + \alpha\operatorname{MSE}(\hat{y}, y) ,
\end{equation}
\noindent
where $\alpha$ denotes the weight of the MSE loss, and $\ell_{\mathrm{MR}}$ is the multi-resolution STFT loss (MR loss).
\vspace{-0.4cm}

\section{Experiments}
\label{sec:experiments}
\vspace{-0.4cm}
\subsection{Setup}
\vspace{-0.1cm}
We focus on extending 4-kHz bandwidth (8 kHz sampling rate) to 8-kHz bandwidth (16 kHz sampling rate). Training data was segmented into smaller chunks with a window size of 8192 and 50\% overlapping. We used the VCTK Corpus \cite{Veaux2017CSTRVC} for training and testing. This dataset includes 109 English speakers, in which recordings of the first 100 speakers were for training and the remaining for testing.

Besides VCTK, we further used the VIVOS dataset \cite{vivos} to verify the effectiveness of our pretraining approach. This dataset consists of 15-hour speech recordings from 65 Vietnamese speakers, recorded in a quiet environment with high-quality microphones. We followed the dataset's default split: 46 speakers for training, 19 speakers for testing. 

To evaluate the quality of the generated audio, we used four metrics: log-spectral distance (LSD), high-frequency log-spectral distance (LSD-HF), scale-invariant source-to-distortion ratio (SI-SDR) \cite{LeRoux2019SDRH}, and DNSMOS based on P.808 criterion \cite{dnsmos}. LSD-HF computes LSD specifically on high-frequency bands, i.e., 4kHz - 8 kHz. As opposed to LSD, LSD-HF focuses only on the regeneration of the high-band spectrum and ignores artifacts or distortions in the low-band spectrum. A lower LSD/LSD-HF score implies a more similar spectral to the target, while a higher SI-SDR score indicates better performance. On the other hand, DNSMOS employs a deep learning model to predict the mean-opinion-score (MOS) of human raters. It has been shown to have excellent correlation to MOS \cite{dnsmos}. A higher value of DNSMOS indicates better speech quality.


The $C$, $K$, $S$, and $B$ hyperparameters of our model are described in Fig.~\ref{fig:unet}. The Performers\footnote{https://github.com/lucidrains/performer-pytorch} block has three hidden layers, two attention heads for each layer, and each head's dimension is 32; local window length is equivalent to bottleneck length divided by 8. Hyperparameters of MR loss such as resolutions were set with default values of the \textit{auraloss}\footnote{https://github.com/csteinmetz1/auraloss} v2.0.1 library. The MSE weight was set to $\alpha=10000$. We trained our models for 150 epochs using the Adam optimizer, $3\times 10^{-4}$ learning rate with 800 samples in each batch. For the baseline TFiLM-UNet model, while official implementation is available, we adopted an unofficial implementation\footnote{github.com/leolya/Audio-Super-Resolution-Tensorflow2.0-TFiLM} which reportedly produces slightly better results and much faster training. 
\vspace{-0.3cm}

\subsection{Performance comparison with baselines}
\vspace{-0.2cm}
\label{sec:per}
We compared our model's performance and inference speed with TFiLM-UNet and two recent generative models, NU-Wave \cite{nuwave} and WSRGlow \cite{wsrglow}. The above baselines were trained on the VCTK dataset with low-rate data generated from 16-kHz data using only one 8th order Chebyshev Type I low-pass filter. In this experiment, MSM pretraining was excluded from our method.
\begin{figure}
\centering
\begin{minipage}[b]{.48\linewidth}
  \raggedright
  \centerline{\includegraphics[width=8.0cm]{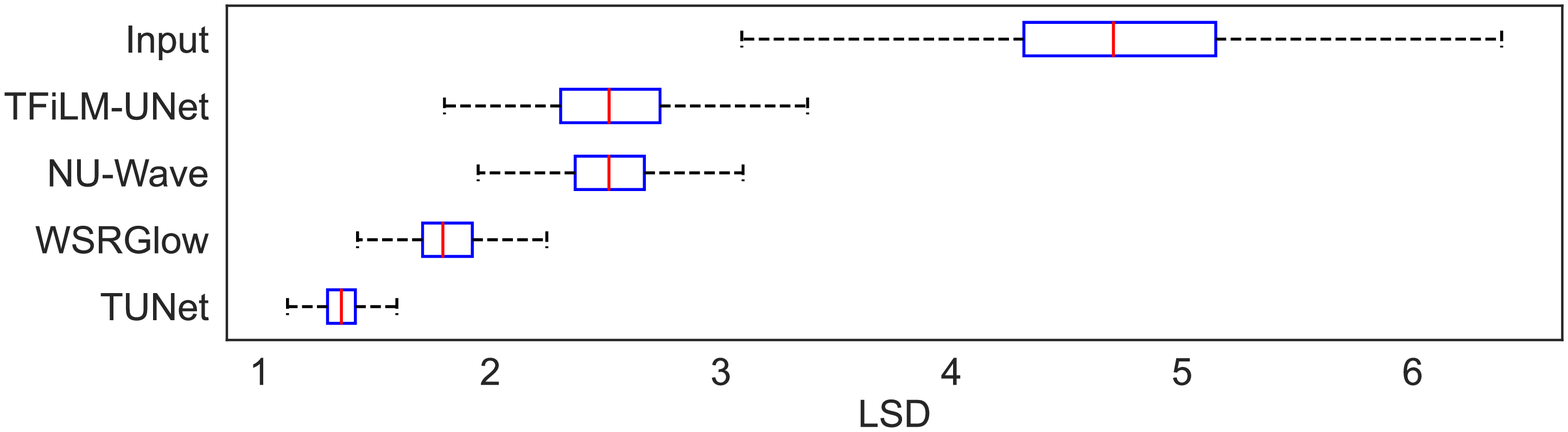}}
\end{minipage}

\begin{minipage}[b]{.48\linewidth}
  \raggedright
  \centerline{\includegraphics[width=8.0cm]{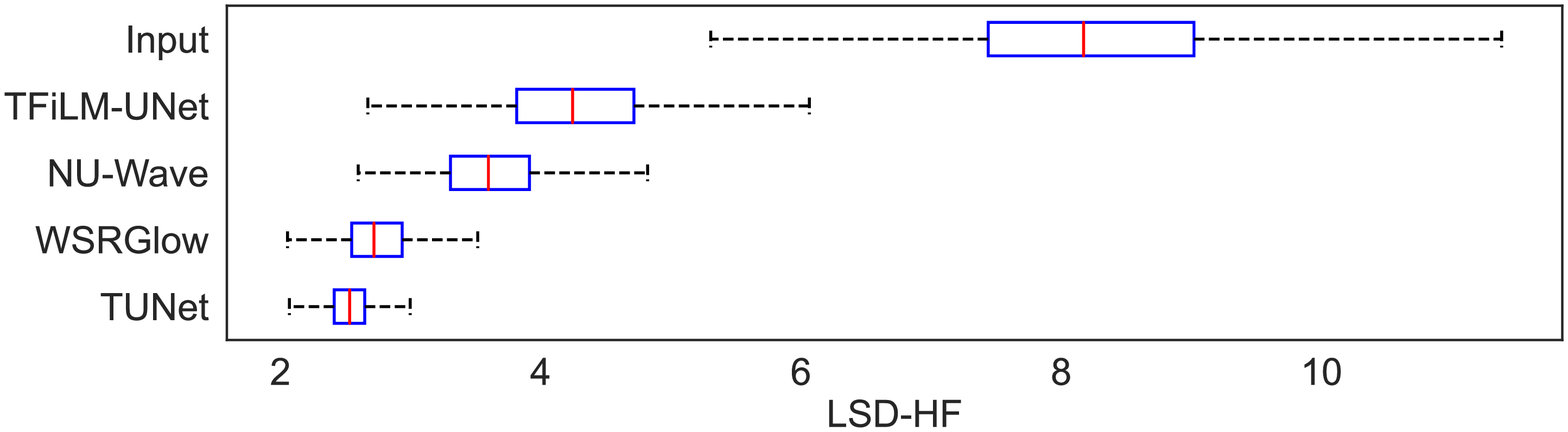}}
\end{minipage}


\begin{minipage}[b]{.48\linewidth}
  \raggedright
  \centerline{\includegraphics[width=8.0cm]{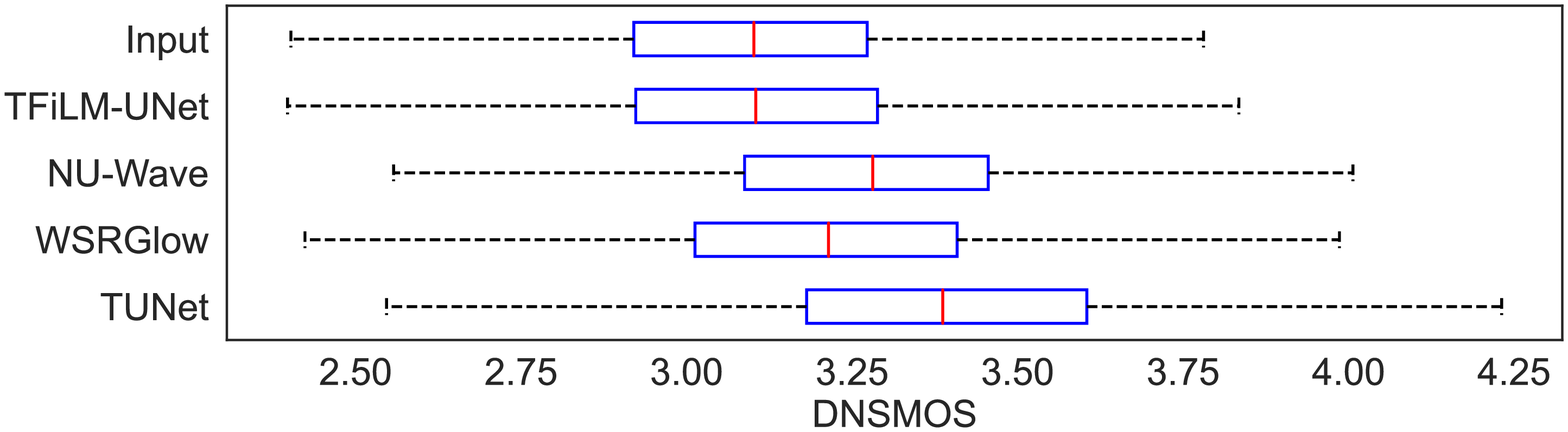}}
\end{minipage}
%
\caption{Metric scores of the baselines and our model. Lower LSD/ LSD-HF is better, and higher DNSMOS is better.}
\label{fig:scores}

\end{figure}

Results in Fig.~\ref{fig:scores} show that our TUNet model achieved significantly higher performance than that of all the baselines.
Compared to our TUNet, the WSRGlow model achieves tight LSD-HF scores but relatively worse in LSD, indicating that our model better preserves low frequencies. Despite the worst LSD score, the NU-Wave model achieves a considerable improvement in DNSMOS only after our proposed model.
\begin{table}
\caption{Model size and inference time on a single core CPU.}
\centering
\resizebox{4.7cm}{!}{%
\begin{tabular}{@{}lll@{}}
\toprule
System     & \#Params      & \multicolumn{1}{c}{\begin{tabular}[c]{@{}c@{}}Inference time \\ (ms)\end{tabular}} \\ \midrule
WSRGlow    & 229M          & 3146                                                                               \\
NU-Wave    & 3M            & 2431 (8 iters)  \\

TFiLM-UNet & 68.2M         & 1335                                                                               \\
TUNet      & \textbf{2.9M} & \textbf{22.6}                                                                      \\ \bottomrule
\end{tabular}%
}

\label{tab:time}
\end{table}

In terms of single-threaded inference time, we measured it on the AMD EPYC 7742 using ONNX inference engine. Our proposed model was significantly faster and more lightweight than the others. In Table~\ref{tab:time}, TUNet requires only 22.63 ms to execute a single 512 ms audio frame while WSRGlow, NU-Wave (the default eight inference steps) and TFiLM-UNet took approximately 139, 107, and 59 times longer, respectively. Assuming each audio chunk being 87.5\% overlapped, this amounts to 64 ms for a new block to arrive with a chunk size of 8192 and a sampling rate of 16 kHz.  Since our inference time is shorter than 64 ms, this implies that the proposed method is more suited for semi-real-time applications compared to the baselines. 
\vspace{-0.35cm}

\subsection{Ablation studies}
\vspace{-0.2cm}
To study the effects of its two main components, TFiLM layers and Performers blocks, we created three variations from TUNet: \textit{`No Transformer'} --- TUNet without Performers blocks on the bottleneck, \textit{`LSTMs bottleneck'} --- TUNet with the Transformer bottleneck replaced by a 3-layer, 256-unit (same as the Transformer) LSTM network, and \textit{`No TFiLM'} --- TUNet without TFiLM layers.

In Table~\ref{tab:comp}, both Performer and TFiLM layers play significant roles in the proposed model since excluding these two components led to noticeably decreased scores on all metrics. The `No Transformer' model, which excluded the Transformer from the bottleneck, performed worst in terms of LSD and SI-SDR, and the performance was only improved by a small margin even with LSTMs aided. The removal of TFiLM also led to a significant degradation but relatively less than the removal of the Transformer.

\begin{table}
\centering
\caption{Effectiveness of components on our model.}
\label{tab:comp}
\resizebox{5.5cm}{!}{%
\begin{tabular}{@{}llll@{}}
\toprule
Model            & \multicolumn{1}{c}{LSD} & \multicolumn{1}{c}{LSD-HF} & \multicolumn{1}{c}{SI-SDR} \\ \midrule
No Transformer   & 1.45                    & 2.64                       & 21.61                      \\
LSTMs bottleneck & 1.44                    & 2.70                       & 21.76                      \\
No TFiLM         & 1.44                    & 2.69                       & 21.89                      \\
TUNet            & \textbf{1.36}           & \textbf{2.54}              & \textbf{21.91}             \\ \bottomrule
\end{tabular}%
}
\end{table}


To determine the effectiveness of MSM pretraining, we pretrained TUNet on VCTK low-rate data with the pipeline described in Section~\ref{sec:msm}. After obtaining a pretrained model, we subsequently trained it with the BWE task on the VCTK dataset. In this experiment, we used only one anti-aliasing filter in Section~\ref{sec:per} to generate training data. To assess the generalization ability of MSM, we include an additional scenario where the pretraining dataset is VCTK, but the BWE training and test set are of a different language. We adopted one more metric --- low-frequency log-spectral distance (LSD-LF) to measure the approximation error in the low band (0-4 kHz) caused by MSM pretraining.

Results in Table~\ref{tab:pretrain} show that models pretrained with MSM achieve significant improvements on spectral-based metrics while SI-SDR figures were modest. The scores indicate that the pretraining scheme not only enhanced high frequencies but also helped preserve low frequencies. Furthermore, the performance gain on the VIVOS was consistent with that of the VCTK. This implies that the BWE model adapted very well to the VIVOS dataset even though it was pretrained on a different language.
\begin{table}
\centering
\caption{BWE results on VCTK and VIVOS datasets when employing MSM pretraining.}
\resizebox{190pt}{!}{%
\begin{tabular}{@{}clllll@{}}
\toprule
\multicolumn{1}{l}{} & Model                                                  & LSD      & LSD-HF        & LSD-LF        & SI-SDR          \\ \midrule
\multirow{3}{*}{VCTK}       & input                                                  & 4.75          & 8.27          & 1.23          & 20.32          \\ 
                            & w/o MSM                                                & 1.36          & 2.54          & 0.18          & 21.69          \\
                            
                            & MSM on VCTK & \textbf{1.28} & \textbf{2.45} & \textbf{0.11} & \textbf{22.08} \\ \midrule
\multirow{3}{*}{VIVOS}      & input                                                  & 5.59          & 9.79          & 1.39          & 21.75          \\ 
                            & w/o MSM                                                & 1.36          & 2.49          & 0.23          & 25.08          \\ 
                            
                            & MSM on VCTK & \textbf{1.29} & \textbf{2.42} & \textbf{0.16} & \textbf{26.15} \\ \bottomrule
\end{tabular}%
}
\label{tab:pretrain}
\end{table}


We next assessed sensitiveness to anti-aliasing filters of our models trained with and without filter augmentation. The first model, `Single Cheby' is the best model obtained from the above experiments, which was trained with a single Chebyshev Type I anti-aliasing filter. The other `Multi-Cheby' was trained with a set of random filters as described in Section~\ref{sec:mfd}. Both models employed the same MSM pretraining above. The BWE dataset used for this experiment was the VIVOS dataset. The test set was downsampled using all resampling methods available in the \textit{resampy}\footnote{https://github.com/bmcfee/resampy} library. However, due to space constraints, we will only report the results on test sets generated by single/multiple Chebyshev filters (same as training of `Single Cheby' and `Multi-Cheby', respectively), Kaiser (`best' and `fast' variations) filters, and the sinc downsampling.

\begin{figure}
    \centering
    \includegraphics[height=2.3cm, width=7.3cm]{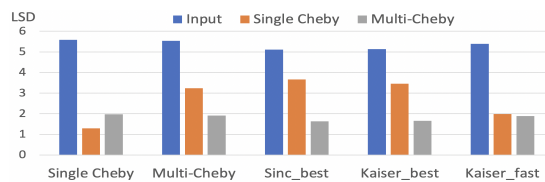}
    \caption{LSD scores of our models trained with a single and multiple anti-aliasing filter(s) on the VIVOS test set.}
    \label{fig:lpf_score}
\end{figure}

As shown in Fig.~\ref{fig:lpf_score}, the `Single Cheby' model achieved the best score when evaluated with the same filter. Although this model performed well on several downsampling methods such as `kaiser\_fast', its performance significantly degraded on test sets processed by the other downsampling methods such as the sinc algorithm. On the other hand, the `Multi-Cheby' showed a stable performance across all the methods. \vspace{-0.6cm}

\section{Conclusions}
\label{sec:conclude}
\vspace{-0.3cm}
We have proposed a Transformer-aided UNet for bandwidth extension. Despite remarkable performance scores, our model remains lightweight and achieves fast processing. By leveraging only narrowband audio data for pretraining, we have achieved an overall improvement in performance. With multiple anti-aliasing filters applied, the model achieves robustness to different low-pass filters, an essential characteristic for real-world applications.
\vfill\pagebreak

\bibliographystyle{IEEEbib}
\bibliography{IEEEabrv, refs}

\end{document}